\documentclass[runningheads]{llncs}

\usepackage{eccv}

\usepackage{eccvabbrv}

\usepackage{graphicx}
\usepackage{booktabs}
\usepackage{amsmath}
\usepackage{amssymb}
\usepackage{mathtools}

\usepackage{algorithm}
\usepackage{algpseudocode}
\usepackage{multirow}
\usepackage{longtable}

\usepackage[accsupp]{axessibility}  

\usepackage{hyperref}

\usepackage{orcidlink}

\begin{document}


\title{Class-wise Contribution Estimation via Logit Maximization for Federated Learning}

\titlerunning{Class-wise Contribution Estimation via Logit Maximization for FL}

\author{Asim Ukaye\inst{1}\orcidlink{0000-0001-6608-8947} \and
  Nurbek Tastan \inst{1}\orcidlink{0000-0002-8182-9819} \and
Mubarak Abdu-Aguye\inst{1}\orcidlink{0000-0001-7010-8103}\and
Karthik Nandakumar\inst{1,2}\orcidlink{0000-0002-6274-9725}
}

\authorrunning{A. Ukaye et al.}

\institute{Mohamed bin Zayed University of Artificial Intelligence, UAE \and
  Michigan State University, USA\\
}

\maketitle

\begin{abstract}
Federated learning (FL) enables collaborative learning of computer vision models, where privacy and regulatory constraints prevent centralizing data across devices or organizations. 
However, practical FL deployments often exhibit severe class imbalance and label skew, causing standard aggregation protocols to overfit dominant clients and degrade minority-class performance. 
We propose a data-free, class-wise \underline{c}ontribution \underline{e}stimation and aggregation framework based on \underline{l}ogit \underline{m}aximization (CELM) that does not require raw data, client metadata, or auxiliary public datasets at the server. 
The FL server probes client updates to obtain class-wise evidence scores and assembles a cross-client evidence matrix, which quantifies both per-class competence and class coverage. 
Using this matrix, we compute contribution weights that upweight clients providing discriminative evidence for underrepresented classes.
The resulting aggregation is stable due to simplex constraints and momentum smoothing, and remains compatible with standard FL training pipelines. 
We evaluate the approach on representative vision benchmarks under controlled non-IID and pathological label splits, demonstrating that CELM-based aggregation improves robustness to imbalance and statistical heterogeneity, while yielding better performance without requiring any additional data exchange.  
The code is available at: \href{https://github.com/asimukaye/celm}{\color{blue} https://github.com/asimukaye/celm}. 


\end{abstract}

\section{Introduction}
\label{sec:intro}

Federated learning (FL) enables collaborative model training across distributed clients without sharing raw data \cite{mcmahanCommunicationEfficientLearningDeep2017}. 
This paradigm is especially attractive for computer vision and machine learning systems deployed across organizations, edge devices, and geographically distributed silos \cite{zhuang2024coala}. 
However, real-world FL systems rarely exhibit independent and identically distributed (IID) client data. 
Clients often differ in dataset sizes, class support, label frequency, and feature quality \cite{li2020federated, Li2020On, tastan2025fedpews, wang2020tackling}. 
Under such heterogeneity, standard aggregation can over-emphasize dominant clients and under-represent clients with minority classes, leading to suboptimal performance across all classes \cite{kairouzAdvancesOpenProblems2021}.

\vspace{.5em}
\noindent A core bottleneck is how to estimate client contribution without relying on self-reported metadata or server-side validation data. 
Metadata-driven weighting can be unreliable or strategically manipulated, while validation-based scoring may be unavailable, biased, or costly to maintain. 
Recent data-free methods partially address this issue, but many are still anchored to similarity-to-average assumptions in gradient/update space, which can undervalue informative clients that carry rare classes or atypical yet useful signals.

\begin{figure*}[t]
  \centering
  \includegraphics[width=\linewidth]{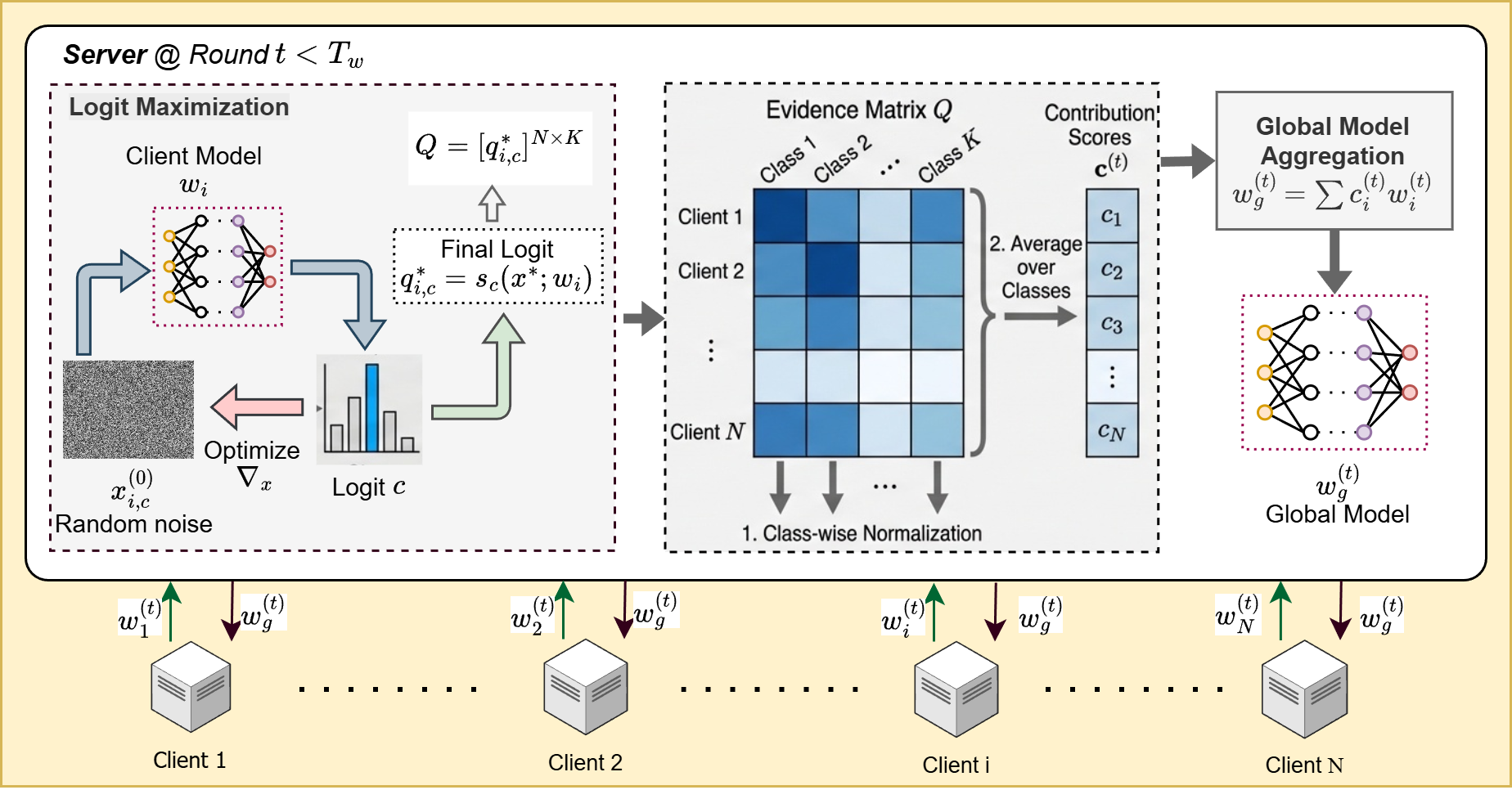}
  \caption{Overview of CELM. During the initial warm-up communication rounds, the server probes each client update using class-wise logit maximization, constructs a debiased client-class evidence matrix, and converts normalized class-wise evidence into contribution weights for aggregation. 
  After warm-up, these contribution weights are frozen and reused for the remaining rounds to stabilize training and reduce overhead.}
  \label{fig:main}
  \vspace{-1em} 
\end{figure*}

\noindent In this work, we propose a \emph{data-free, class-wise contribution estimation} mechanism based on \emph{logit maximization}. 
During an initial warm-up window, the server probes each client model class-by-class by optimizing noisy inputs to maximize class logits. 
These probes yield a client-class evidence matrix that captures relative per-class evidence across clients. 
We then de-bias evidence using a global reference, normalize class-wise shares, and aggregate them into client contribution scores that are then used for weighted model aggregation.

\noindent This design offers three main advantages. 
\textbf{First,} it does not require any auxiliary validation set or client-reported metadata at the server. 
\textbf{Second,} class-wise normalization improves robustness under label skew by valuing relative class share rather than dominant-class overlap. 
\textbf{Third,} estimation during warm-up followed by score freezing and EMA smoothing yields stable aggregation with bounded server computational overhead.
These properties make the method effective not only for standard non-IID splits, but also for extreme cases such as Maverick clients (informative rare-class holders) and for detecting free-riders (uninformative contributors or non-contributors).

\noindent We evaluate the method on representative vision benchmarks, including FashionMNIST, CIFAR-10, and FedISIC, under controlled non-IID partitions and stress-test settings. 
Beyond predictive performance under data heterogeneity, we analyze three additional aspects: 
(i) performance with rare-class holding clients (Mavericks), 
(ii) free-rider detection from contribution estimates, and 
(iii) fidelity of estimated client/global class distributions compared to the true distribution.
Together, these studies show that logit-maximization-based scoring is both useful for aggregation and semantically meaningful as a proxy for client distributional contribution.
The key contributions of this paper are:
\begin{itemize}
    \setlength\itemsep{.2em}
  \item We introduce a data-free, class-wise contribution estimator for FL based on server-side logit maximization.
  \item We propose a class-share-based scoring rule that captures both informative rare-class clients and uninformative non-contributing clients.
  \item We design a warm-up-and-freeze aggregation strategy with EMA smoothing that improves stability and limits overhead.
  \item We provide extensive empirical evidence on global performance across non-IID splits and under Maverick and free-rider scenarios. We also provide further insights through a distribution estimation fidelity analysis.
\end{itemize}

\section{Related Work}
\label{sec:rel}

\paragraph{\textbf{Contribution Estimation in Federated Learning.}}
A central challenge in FL is to fairly weight or reward clients based on their actual contribution to the global model. 
Early work, such as FedAvg \cite{mcmahanCommunicationEfficientLearningDeep2017}, uses the number of local samples as a proxy for client weight, implicitly assuming honest self-reporting. 
However, when incentives are tied to contribution, this assumption may not hold \cite{tastan2025aequa}.
More recent approaches rely on an auxiliary validation set to assess each client’s update. 
For example, CFFL \cite{lyuCollaborativeFairnessFederated2020} and FedCE \cite{jiang2023fair} evaluate client models on server-side validation data to estimate their goodness. 
However, validation sets may be unavailable or may be biased relative to the actual client data distributions, limiting the reliability of these metrics \cite{Li2020On, li2020federated, karimireddy2020scaffold, tastan2025fedpews, li2021moon}. 
Several works provide a detailed taxonomy and a comprehensive review of collaborative fairness and robustness in federated learning \cite{shi2023towards, jaramillo2026private, huang2024benchmark}. 

\paragraph{\textbf{Validation Data-Free and Non-Self-Reported Metadata Methods.}}
To avoid privacy risks or reliance on client metadata, several works propose validation data-free scoring of client updates \cite{xuGradientDrivenRewards2021, tastan2024redefining, tastan2025cycle, tastan2025aequa}. 
CGSV \cite{xuGradientDrivenRewards2021} computes the cosine similarity between the gradient of each client and the average gradient of the cohort, rewarding updates that are better aligned with the average gradient. 
ShapFed \cite{tastan2024redefining} improves upon the gradient alignment strategy by proposing class-wise gradient alignment for the final layer of the model. 
This allows for better estimates of client contributions in scenarios with a high skew in class labels within each client. 
These methods highlight a trend in server-side evaluation of client updates without using any auxiliary data. 

\paragraph{\textbf{Data Heterogeneity and Label Skew in FL.}}
Several FL methods address non-IID training through client-side optimization or classifier corrections.
FedProx~\cite{li2020federated}, SCAFFOLD~\cite{karimireddy2020scaffold}, FedNOVA~\cite{wang2020tackling}, and FedDyn~\cite{acar2021federated} reduce client drift or objective inconsistency under heterogeneous local data. 
MOON~\cite{li2021moon} uses representation alignment to reduce drift.
Label-skew-specific methods include FedRS~\cite{li2021fedrs}, which restricts softmax computation to mitigate missing-label bias.
FedUV~\cite{son2024feduv} regularizes classifier uniformity and variance to address label skew.
Other client-side methods address adjacent heterogeneity settings through sharpness aware optimization in FedSAM~\cite{qu2022generalized}, stabilized orthogonal/proximal learning in FedSOL~\cite{lee2024fedsol}, learning from positive labels only in FedAwS~\cite{yu2020federated}, and enforcing ETF geometry in FedGELA~\cite{fan2023federated}.
These approaches primarily alter client-side training objectives or classifier behavior.

\paragraph{\textbf{Free-riders and Mavericks in FL.}}
Recent FL works highlight two challenging client types under heterogeneity. \emph{Mavericks} \cite{huang2023maverick, huang2022tackling, yangrewarding} are clients that may hold rare but highly informative classes. 
\emph{Free-riders} \cite{he2024client, fraboni2021free } contribute poor or non-informative updates while still benefiting from global training. 
Similarity-to-average weighting, such as in CGSV, can suppress Maverick clients because their updates are atypical. 
On the other hand, naive sample-count or uniform weighting can over-credit free-riders. 
Prior fairness and reputation-oriented methods \cite{lyuCollaborativeFairnessFederated2020, xuGradientDrivenRewards2021, Xu2021RFFL, tastan2025cycle, tastan2025aequa} address parts of this issue by differentiating client rewards from contribution scoring or reducing the impact of low-quality contributors.
However, most operate primarily in the update space rather than class-wise evidence space. 
Our method complements this line by estimating class-wise evidence directly from client models, which helps preserve rare-class signal and suppress non-informative clients during aggregation.

\paragraph{\textbf{Logit/Activation Maximization.}}
Activation maximization (AM) and logit maximization techniques have been used to probe class-selective behavior and internal representations of deep networks by optimizing inputs to increase a target neuron or class logit \cite{erhan2009visualizing, simonyan2014visualising, yosinski2015understanding}. 
Subsequent work emphasized practical priors and optimization choices to improve interpretability and reduce high-frequency artifacts, including feature-visualization refinements \cite{olah2017feature} and smoothness-inducing regularization \cite{mahendran2016visualizing}. 
In contrast to interpretability use cases, our method repurposes class-wise logit maximization as a server-side probing mechanism in federated learning. 
To the best of our knowledge, our method is the first use of logit maximization for contribution scoring in federated learning.

\vspace{.5em}
\section{Preliminaries}
\label{sec:prelim}

\subsection{Federated Learning Setup}
\label{sec:flsetup}

We consider a cross-silo federated learning (FL) system with $N$ clients indexed by $i\in\mathcal{N}$, where $\mathcal{N}=\{1,\dots,N\}$. 
Each client $i$ has a private dataset $\mathcal{D}_i = \{(\mathbf{x}_{i,j},y_{i,j})\}_{j=1}^{n_i}$, where $\mathbf{x}_{i,j} \in \mathcal{X}$ denotes an image with class label $y_{i,j} \in \mathcal{Y}$.
$\mathcal{X}$ represents the image space, $\mathcal{Y}=\{1,\dots,K\}$ represents the label space.
$K$ is the number of classes, and $n_i$ is the size of the training set of client $i$. 
The overarching goal of the system is to enable clients to collaboratively learn a multi-class classifier $g_w: \mathcal{X} \rightarrow \mathcal{Y}$ without sharing their private data, where
$w \in \mathbb{R}^{d}$ denotes the model parameters. 
Let $F_i(w;(\mathbf{x},y)) = \mathcal{L}_i(g_w(\mathbf{x}),y)$ be the per-sample loss function at client $i$. The global optimization objective of the FL system is:
\begin{align}
  \min_{w\in\mathbb{R}^d} f(w), \quad 
  f(w)=\frac{1}{N}\sum_{i=1}^{N} f_i(w), \quad 
  f_i(w)=\mathbb{E}_{(\mathbf{x},y)\sim\mathcal{D}_i}\big[F_i(w;(\mathbf{x},y))\big].
  \label{eq:fl-objective}
\end{align}
The server initializes the global model parameters $w_g^{(0)}$. 
In communication round $t \in [1,\dots,T]$, the server broadcasts the global parameters $w_g^{(t-1)}$. 
Each client performs local optimization based on this initialization and returns updated weights $w_i^{(t)}$. 
The server then aggregates the client models using client-specific weights $\mathbf{c}^{(t)}=[c_1^{(t)},\dots,c_N^{(t)}]^\top$:
\begin{align}
  w_g^{(t)} = \sum_{i=1}^{N} c_i^{(t)} w_i^{(t)},
  \qquad c_i^{(t)}\ge 0, \qquad \sum_{i=1}^{N} c_i^{(t)}=1.
  \label{eq:weighted-agg}
\end{align}
In the classical FedAvg algorithm \cite{mcmahanCommunicationEfficientLearningDeep2017}, clients are assigned weights based on their self-reported sample sizes, i.e., $c_i^{(t)} = \frac{n_i}{\sum_{j=1}^{N}n_j}$. 
Unlike FedAvg, we aim to estimate $\mathbf{c}^{(t)}$ from class-wise evidence obtained by server-side probing of the client models, without using any validation data or client self-reported metadata. 

\subsection{Activation Maximization}

Activation maximization (AM) is a well-known approach \cite{erhan2009visualizing} that attempts to find an input that maximizes a target functional $\mathcal{T}(\mathbf{x})$, such as a neuron, under a regularization prior $\mathcal{R}(\mathbf{x})$:
\begin{align}
  \mathbf{x}^{\star}=\arg\max_{\mathbf{x}\in\mathcal{X}}\; \mathcal{T}(\mathbf{x})-\lambda \mathcal{R}(\mathbf{x}),
  \label{eq:am-general}
\end{align}
where $\lambda\geq 0$ controls the activation-regularization trade-off. 
Previous works in the AM literature have used total variation (TV) regularization to suppress high-frequency artifacts, especially for image inputs \cite{mahendran2016visualizing, yosinski2015understanding}.

\section{Proposed Methodology}
\label{sec:method}

We present a data-free contribution assessment method for federated learning called \textit{\underline{C}ontribution \underline{E}stimation from \underline{L}ogit \underline{M}aximization} (CELM). 
The method has three stages: 
(i) extract class-specific evidence with logit maximization, 
(ii) convert that evidence into stable aggregation weights, and 
(iii) reuse frozen weights after warm-up to reduce overhead. 

\paragraph{\textbf{Scope and assumptions.}}
We focus on the \textit{cross-silo} FL setting with a small to moderate number of clients whose data may differ in sample size and label distribution.
There is no auxiliary validation data or client-reported metadata available at the server.
Here, metadata denotes side information beyond the required FL model update that is reported by the clients, such as dataset size, class histograms, or validation losses/scores \cite{shi2023towards}. 
Although self-reported metadata is not inherently unreliable, contribution scoring and reward distribution mechanisms based on it may incentivize clients to manipulate it.  
We assume an honest-but-curious server that may inspect received client updates for contribution scoring or free-rider detection \cite{boenisch2023curious}.
We do not address adversarial or Byzantine clients that may actively try to corrupt the FL process.

\subsection{Logit Maximization Framework}

In this work, we utilize the class-specific logit activation as the target functional in AM. 
Let $s_c(\mathbf{x};w)$ be the pre-softmax logit for class $c \in \mathcal{Y}$ output by the classifier $g_w$ based on input $\mathbf{x}$. 
Given only a classification model $g_w$, the logit maximization (LM) framework attempts to find an input $\mathbf{x}$ that maximizes the activation of the class-specific logits, i.e., $\mathcal{T}(\mathbf{x}) = s_c(\mathbf{x};w)$. 
For a given model $g_{w}$ and class $c$, we solve the following problem.

\begin{align}
  \mathbf{x}_{c}^{\star}=\arg\max_{\mathbf{x}\in\mathcal{X}}\; s_{c}(\mathbf{x};w)-\lambda\|\mathbf{x}\|_2^2.
  \label{eq:logit-max}
\end{align}

\noindent Starting from a random initialization for $\mathbf{x}$, we optimize the above objective via gradient ascent. 
For simplicity, we use only $\ell_2$-norm based regularization, i.e., $\mathcal{R}(\mathbf{x})=\|\mathbf{x}\|_2^2$, to limit unbounded pixel growth and keep optimization stable. 

\noindent Let $q_{c}$ be the final logit value after a fixed number of optimization steps.
\begin{align}
  q_{c}=s_{c}(\mathbf{x}_{c}^{\star};w).
  \label{eq:probe-score}
\end{align}
\noindent This optimized logit score $q_{c}$ can be used as a proxy for the class-specific evidence. 
The intuition behind this choice is as follows. If the model $g_w$ is trained on sufficient samples from a specific class $c$, it can be expected to learn the attributes of this class well. 
Consequently, when probed using the LM framework, it should be possible to obtain an input that results in a high logit value $q_{c}$. 
In contrast, if the model $g_w$ is trained on a negligible number of samples from class $c$, it cannot learn the characteristics of this class well, and consequently, it will fail to produce a high logit score $q_{c}$ under LM probing.

\subsection{Class-specific Client Contribution Estimation in FL}

Now, consider the FL setup described in \cref{sec:flsetup}. 
In communication round $t$, after receiving client models $\{w_i^{(t)}\}_{i=1}^N$, the server performs class-wise probing for each client model. 
For each client $i \in \mathcal{N}$ and class $c \in \mathcal{Y}$, the server randomly initializes an input $\mathbf{x}$ and runs $L$ gradient ascent steps with step size $\eta$ to obtain:

\begin{align}
  \mathbf{x}_{i,c}^{\star,(t)} = \arg\max_{\mathbf{x}} \; s_{c}(\mathbf{x}; w_i^{(t)})-\lambda\lVert \mathbf{x}\rVert_2^2.
  \label{eq:celm_probe_obj}
\end{align}

\noindent The above objective drives the synthetic input toward class-specific evidence while regularization controls degenerate solutions. We then compute the raw client-class logit as:
\begin{align}
  \tilde{q}_{i,c}^{(t)} = s_{c}(\mathbf{x}_{i,c}^{\star,(t)}; w_i^{(t)}).
  \label{eq:celm_q_raw}
\end{align}
To de-bias this value, we run the same logit-maximization step in \cref{eq:celm_probe_obj} on the global model from the previous round $w_g^{(t-1)}$ for class $c$ to obtain the corresponding optimized input $\mathbf{x}_{g,c}^{\star,(t-1)}$. 
The bias is then estimated as:
\begin{align}
  b^{(t-1)} = \frac{1}{K}\sum_{c=1}^{K} s_{c}\!\left(\mathbf{x}_{g,c}^{\star,(t-1)}; w_g^{(t-1)}\right).
  \label{eq:celm_baseline}
\end{align}
This baseline captures the global model's generic confidence level and helps remove shared calibration effects across clients. 
The final evidence score applies baseline subtraction followed by ReLU suppression:
\begin{align}
  q_{i,c}^{(t)} = \max\!\left(0,\; \tilde{q}_{i,c}^{(t)} - b^{(t-1)}\right).
  \label{eq:celm_q}
\end{align}
Stacking all $q_{i,c}^{(t)}$ forms the evidence matrix $Q^{(t)}\in\mathbb{R}^{N\times K}$. 
In practice, this matrix is the key intermediate object that summarizes class-wise client utility for round $t$. 
Note that each communication round in FL aggregates the client models to obtain a global model, which is then used as the initialization for local training in the next round. 
Therefore, after the initial communication rounds, the global model will acquire knowledge about all the classes as long as one or more clients have sufficient samples of each class.
Consequently, it becomes increasingly difficult to obtain class-specific evidence from the client models. 
Hence, we apply the LM framework only when $t\le T_w$, where $T_w$ is the warm-up horizon. 

\vspace{.5em}
\noindent To obtain relative class share, CELM normalizes per-class evidence across clients:
\begin{align}
  r_{i,c}^{(t)} = \frac{q_{i,c}^{(t)}}{\sum_{j=1}^{N} q_{j,c}^{(t)} + \epsilon},
  \label{eq:celm_rel}
\end{align}
where $\epsilon>0$ ensures numerical stability. This class-wise normalization is important as it prevents globally dominant clients from trivially dominating all classes. 
This normalization is empirically important in the Maverick and label-skew settings studied in Sec.~\ref{sec:results}.
The client contribution score is then the average relative share over classes:
\begin{align}
  \hat{c}_i^{(t)} = \frac{1}{K}\sum_{c=1}^{K} r_{i,c}^{(t)}.
  \label{eq:celm_avg}
\end{align}
This averaging step gives a single interpretable score per client while still preserving class-aware evidence in the computation.

\noindent Next, we compute instantaneous simplex-normalized scores as follows:
\begin{align}
  \bar{c}_i^{(t)} = \frac{\hat{c}_i^{(t)}}{\sum_{j=1}^{N} \hat{c}_j^{(t)}}, \qquad \bar{c}_i^{(t)}\ge 0,\; \sum_{i=1}^{N}\bar{c}_i^{(t)}=1.
  \label{eq:celm_simplex}
\end{align}
Subsequently, we apply exponential moving average (EMA) smoothing with factor $\beta\in[0,1)$:
\begin{align}
  c_i^{(t)} = \beta c_i^{(t-1)} + (1-\beta)\bar{c}_i^{(t)}.
  \label{eq:celm_ema}
\end{align}
EMA dampens the round-to-round volatility in estimated contributions and improves aggregation stability when client updates are noisy.

\noindent CELM computes $\mathbf{c}^{(t)}$ only during the initial $T_w$ communication rounds. 
During this warm-up phase, the server computes the full global model by aggregating the client models, but only shares the global backbone with the clients. 
The clients retain their local classifier layer; empirically, this preserves a stronger class-discriminative signal in the probed logits. 
After the warm-up phase, the final score vector is frozen:
\begin{align}
  \mathbf{c}^{(t)} = \mathbf{c}^{(T_w)}, \qquad \forall~ t > T_w.
  \label{eq:celm_freeze}
\end{align}

After $T_w$, the server broadcasts the full aggregated model (including classifier head) in each subsequent round. 
This design keeps per-round overhead low after early calibration while preserving class-aware contribution signals. 
\cref{alg:celm} in the appendix summarizes the complete procedure.

\vspace{-0.5em}
\section{Experimental Setup}
\vspace{-0.5em}

\begin{figure*}[t]
  \centering
  \begin{minipage}[t]{0.32\linewidth}
    \centering
    \includegraphics[width=\linewidth]{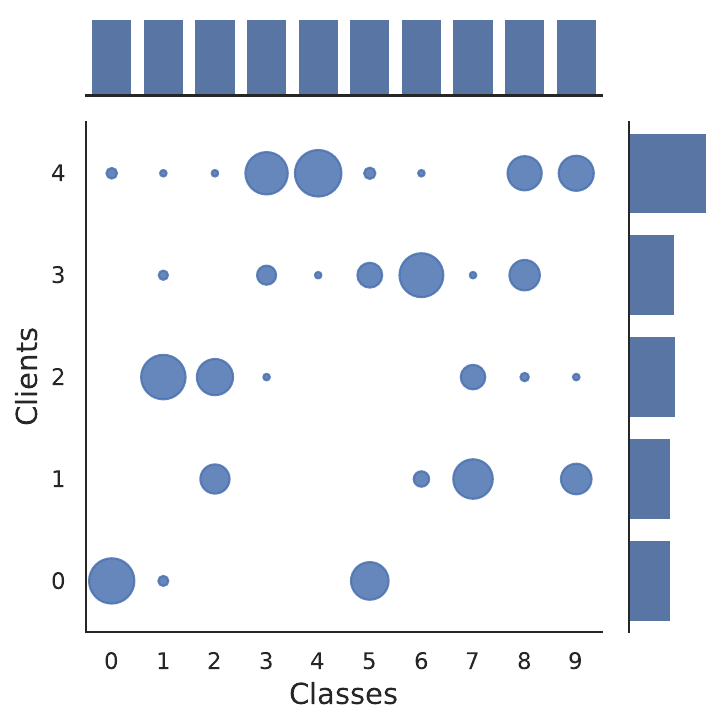}
    \vspace{2pt}
    \small (a) Dirichlet ($\alpha=0.05$)
  \end{minipage}\hfill
  \begin{minipage}[t]{0.32\linewidth}
    \centering
    \includegraphics[width=\linewidth]{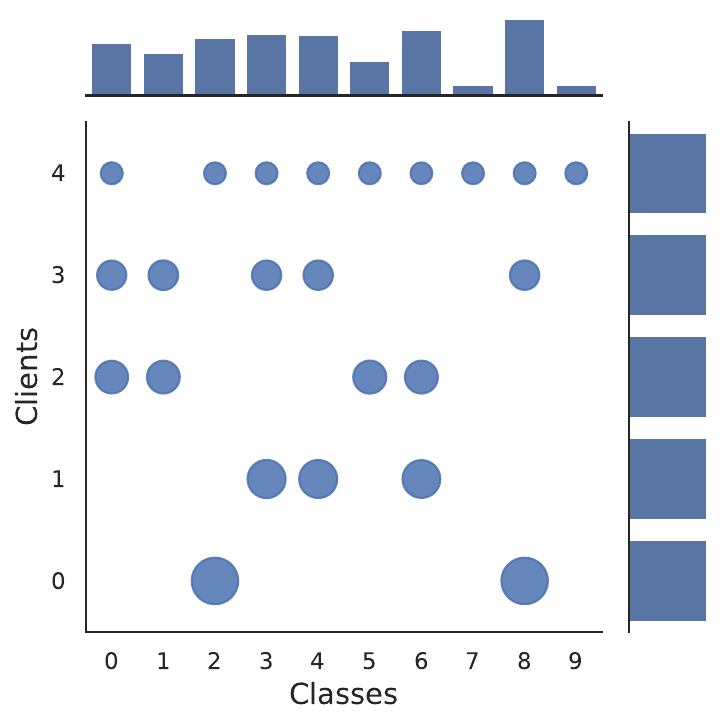}
    \vspace{2pt}
    \small (b) Pure Label Skew (PLS)
  \end{minipage}\hfill
  \begin{minipage}[t]{0.32\linewidth}
    \centering
    \includegraphics[width=\linewidth]{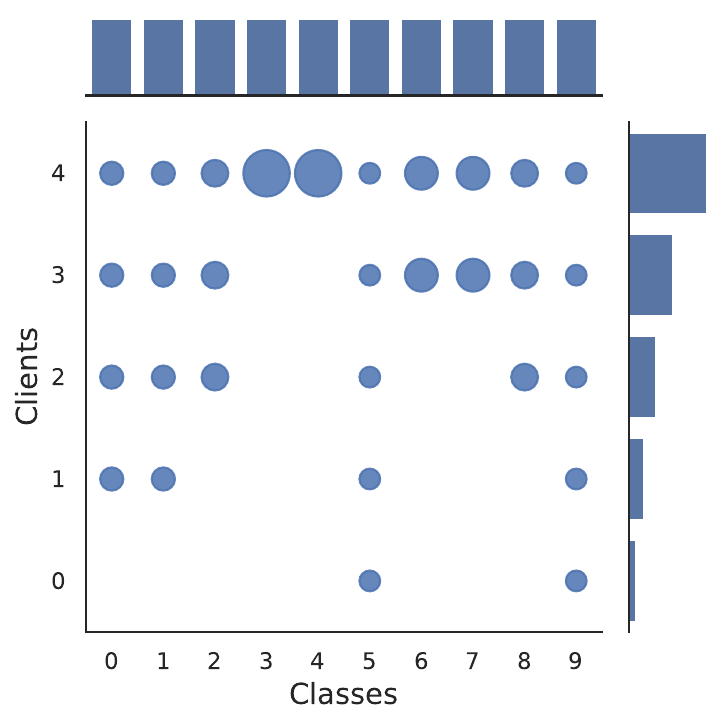}
    \vspace{2pt}
    \small (c) Step Label Skew (SLS)
  \end{minipage}
  \caption{Visualization of the non-IID split settings used in our experiments. 
  Each bubble plot encodes the client-class label allocation. The bubble size is proportional to the number of samples of a given class held by a given client. 
  The top marginal summarizes per-class totals across all clients, while the right marginal summarizes per-client totals across all classes.}
  \label{fig:exp_setting}
  \vspace{-1.2em}
\end{figure*}

\paragraph{\textbf{Datasets and Models.}}
We evaluate CELM on FashionMNIST, CIFAR-10, and FedISIC. For FashionMNIST, we use a 4-layer MLP. For CIFAR-10, we use a randomly initialized 5-layer CNN. 
For FedISIC, we use an ImageNet-1k pretrained \text{ViT-B/16}. 

\paragraph{\textbf{Training Hyperparameters.}}
We simulate $5$ clients for FashionMNIST and CIFAR-10, and $6$ clients for FedISIC.
Each client performs one local epoch per communication round. We run $100$ rounds for FashionMNIST, and $200$ rounds for CIFAR-10 and FedISIC.
We use one local epoch to isolate the effect of the aggregation rule and reduce confounding from client drift; 
varying the number of local epochs is an implementation choice rather than a methodological restriction.
Client-side optimization uses vanilla SGD with cross-entropy loss and batch size $128$. 
The initial learning rate is $0.1$ for FashionMNIST and CIFAR-10, and $0.01$ for FedISIC, with a step schedule that decays the learning rate by a factor of $0.1$ after 50 rounds. 
The hyperparameters chosen are tuned on the FedAvg algorithm to achieve optimal performance with the fewest rounds.

\paragraph{\textbf{LM and Contribution Estimation Hyperparameters.}}
For logit maximization probing, we use Adam with an LM learning rate $0.01$ and $200$ optimization steps per class-client probe. 
The $\ell_2$-regularization coefficient $\lambda$ is set to 0.001. 
The images for LM are initialized with random Gaussian noise from the standard normal distribution for the first round. 
For subsequent rounds, the final image from the previous round is used as the starting point. The EMA factor for contribution smoothing is set to $\beta=0.5$. 
The warm-up horizon is set to $5\%$ of total communication rounds, i.e., $T_w=0.05T$. 
An ablation study on the choice of LM optimization parameters and warm-up horizon is given in Appendix \ref{app:ablation}.

\paragraph{\textbf{Data Splits.}}
For FashionMNIST and CIFAR-10, we consider three non-IID regimes, illustrated in \cref{fig:exp_setting}. 

\begin{itemize}
  \item \textbf{Pure Label Skew (PLS)} assigns different numbers of classes to different clients while keeping the total number of samples per client fixed, isolating the effect of label-support imbalance.
  \item \textbf{Step Label Skew (SLS)} increases both the number of classes and the sample count across clients in a step-wise manner, creating a coupled label-and-quantity heterogeneity pattern.
  \item \textbf{Dirichlet splits} use class-wise Dirichlet sampling with concentration parameter $\alpha \in \{0.01, 0.05, 0.1\}$ to generate randomized non-IID partitions; smaller $\alpha$ yields stronger skew and larger cross-client imbalance.
\end{itemize}
 \noindent For FedISIC, we use the naturally provided client partition.
\paragraph{\textbf{Baselines and comparison protocol.}} 
We primarily compare CELM against FedAvg, CFFL, CGSV, and ShapFed, which together represent the commonly used baselines for contribution-aware FL under heterogeneous data. 
CGSV and ShapFed are validation-free and do not rely on client metadata, making them the most directly comparable baselines to CELM. 
CFFL provides a validation-based baseline. The validation set in CFFL follows the client distribution to reflect the heterogeneity present in the client data.
To keep the comparison compatible with the no client-metadata setting, we report FedAvg with uniform client weighting, so no client-reported metadata is used in aggregation. 
In addition to the data-free contribution estimation baselines, we compare against standard baselines in FL that address label skew and data heterogeneity. 
All performance results are reported as $\text{mean}_{\pm \text{std.}}$ over three random seeds.

\vspace{-0.8em}
\section{Results}
\vspace{-0.5em}
\label{sec:results}

\begin{table*}[t!]

  \centering
  \caption{
  Global predictive performance of the methods across datasets and splits. 
  \textdagger We report the balanced accuracy for FedISIC to account for the label imbalance in its test set. 
  The best results are shown in \textbf{bold}. Second-best results are \underline{underlined}.\label{tab:main_results}
  }

  \begin{tabular}{l l c c c c c}
    \toprule
    {\textbf{Dataset} }& {\textbf{Split}} &{ \textbf{FedAvg}} & {\textbf{CFFL}} & \textbf{CGSV} &  {\textbf{ShapFed}} & \textbf{CELM} \\
    \midrule
    \multirow{5.2}{*}{F.MNIST}
    & Dir. (0.01) & $80.64_{\pm 2.25}$ & $78.89_{\pm 2.93}$ & $47.40_{\pm 7.82}$ & \underline{$81.15_{\pm 1.83}$} & $\mathbf{81.76_{\pm 1.85}}$ \\
    & Dir. (0.05) & $81.63_{\pm 2.69}$ & $81.89_{\pm 0.45}$ & $46.30_{\pm 3.34}$ & \underline{$81.97_{\pm 2.61}$} & $\mathbf{83.64_{\pm 0.42}}$ \\
    & Dir. (0.10) & $84.83_{\pm 0.62}$ & $84.58_{\pm 0.64}$ & $63.37_{\pm 3.71}$ & \underline{$85.14_{\pm 0.65}$} & $\mathbf{85.34_{\pm 0.09}}$ \\
    & PLS & $80.62_{\pm 0.52}$ & $80.95_{\pm 1.81}$ & $49.41_{\pm 2.17}$ & \underline{$81.80_{\pm 0.56}$} & $\mathbf{83.70_{\pm 0.19}}$ \\
    & SLS & $83.13_{\pm 2.61}$ & $\mathbf{88.13_{\pm 0.35}}$ & $55.39_{\pm 5.06}$ & $84.17_{\pm 2.07}$ & \underline{$87.00_{\pm 1.14}$} \\ [0.3em]
    \midrule
    \multirow{5.2}{*}{CIFAR-10}
    & Dir. (0.01) & \underline{$63.41_{\pm 1.42}$} & $53.59_{\pm 6.74}$ & $18.25_{\pm 6.09}$ & $63.12_{\pm 1.39}$ & $\mathbf{64.37_{\pm 0.98}}$ \\
    & Dir. (0.05) & $65.12_{\pm 1.89}$ & $55.92_{\pm 9.31}$ & $24.48_{\pm 1.57}$ & \underline{$65.83_{\pm 1.71}$} & $\mathbf{67.16_{\pm 1.29}}$ \\
    & Dir. (0.10) & $68.98_{\pm 0.67}$ & $60.41_{\pm 4.85}$ & $29.03_{\pm 3.16}$ & \underline{$68.98_{\pm 0.42}$} & $\mathbf{69.21_{\pm 0.76}}$ \\
    & PLS         & \underline{$56.82_{\pm 2.96}$} & $49.93_{\pm 6.28}$ & $28.75_{\pm 4.92}$ & $56.80_{\pm 2.80}$ & $\mathbf{59.11_{\pm 1.96}}$ \\
    & SLS         & $67.40_{\pm 0.95}$ & \underline{$70.47_{\pm 2.10}$} & $33.58_{\pm 1.76}$ & $69.91_{\pm 0.43}$ & $\mathbf{71.96_{\pm 0.08}}$ \\
    [0.3em]
    \midrule
    FedISIC\textdagger & Natural & $61.25_{\pm 0.04}$ & \underline{$62.60_{\pm 6.34}$} & $26.17_{\pm 0.59}$ & $62.31_{\pm 0.16}$ & $\mathbf{70.18_{\pm 0.58}}$ \\
    \bottomrule
  \end{tabular}
\end{table*}

\noindent \cref{tab:main_results} shows that CELM is consistently competitive and often the best across heterogeneous settings, with higher gains under stronger non-IID skew. 
On FashionMNIST, CELM achieves the top score in four of the five reported splits. 
The improvements over FedAvg are most pronounced in challenging skewed regimes, e.g., PLS ($83.70$ vs. $80.62$) and Dirichlet(0.05) ($83.64$ vs. $81.63$). 
On CIFAR-10, CELM is the strongest method in all five reported splits. 
The margins are again largest in highly heterogeneous settings: $+2.29$ points over FedAvg in PLS ($59.11$ vs. $56.82$), +$8.80$ over CFFL in Dirichlet(0.1), and +$2.05$ over ShapFed in SLS. 
These results indicate that class-wise contribution weighting provides meaningful robustness when label support is uneven across clients.

\vspace{.5em}
\noindent On FedISIC (natural split), CELM yields the best balanced accuracy ($70.18$), substantially outperforming the next best method CFFL ($62.60$) by a margin of $+7.58$. 
This large gap on a naturally partitioned clinical dataset supports the practical utility of CELM beyond synthetic splits, where both label-distribution mismatch and client variability are intrinsic.

\vspace{-1em}
\subsection{Performance on clients with unique classes (Mavericks)}
\begin{figure*}[t]
  \centering
  \begin{minipage}[t]{0.32\linewidth}
    \centering
    \includegraphics[width=\linewidth]{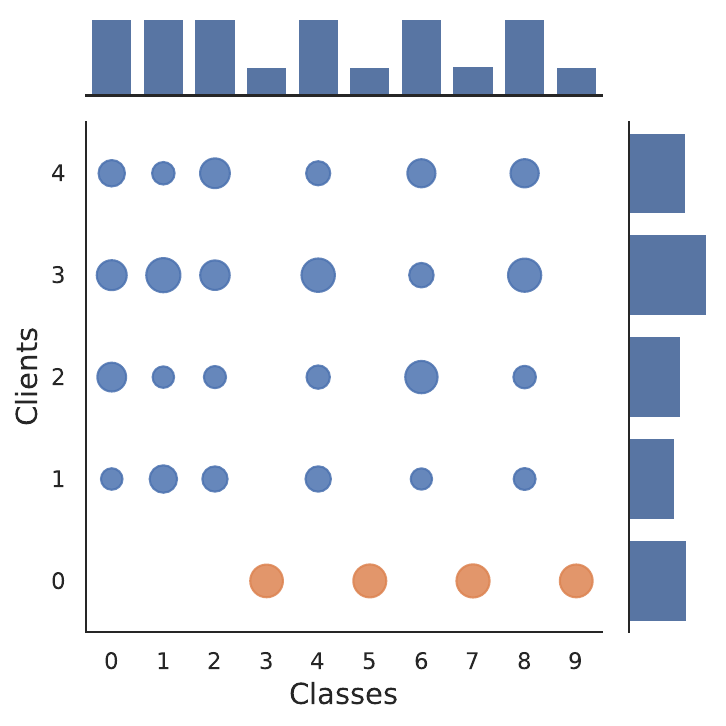}
    \vspace{2pt}
    \small (a) Maverick Client
  \end{minipage}
  \hfill
  \begin{minipage}[t]{0.32\linewidth}
    \centering
    \includegraphics[width=\linewidth]{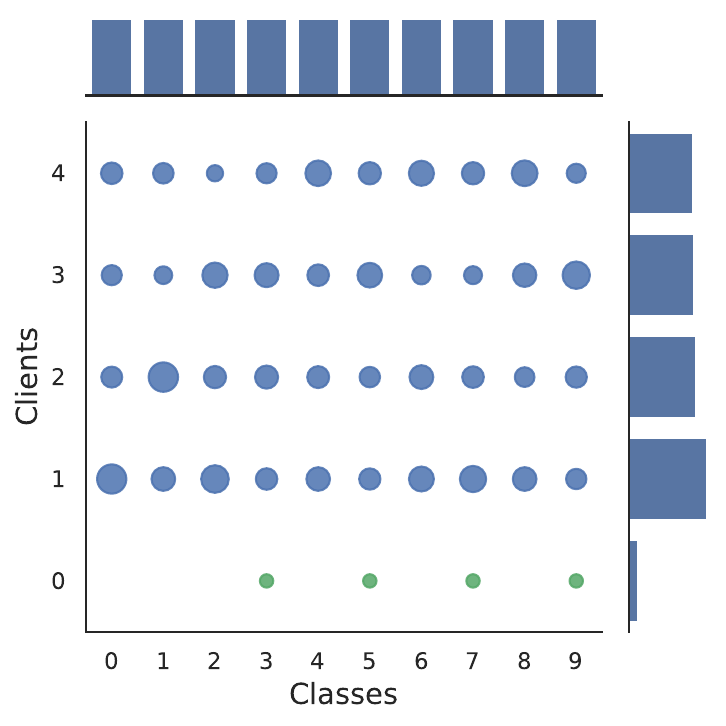}
    \vspace{2pt}
    \small (b) Free-rider Client
  \end{minipage}
  \hfill
  \begin{minipage}[t]{0.32\linewidth}
    \centering
    \includegraphics[width=\linewidth]{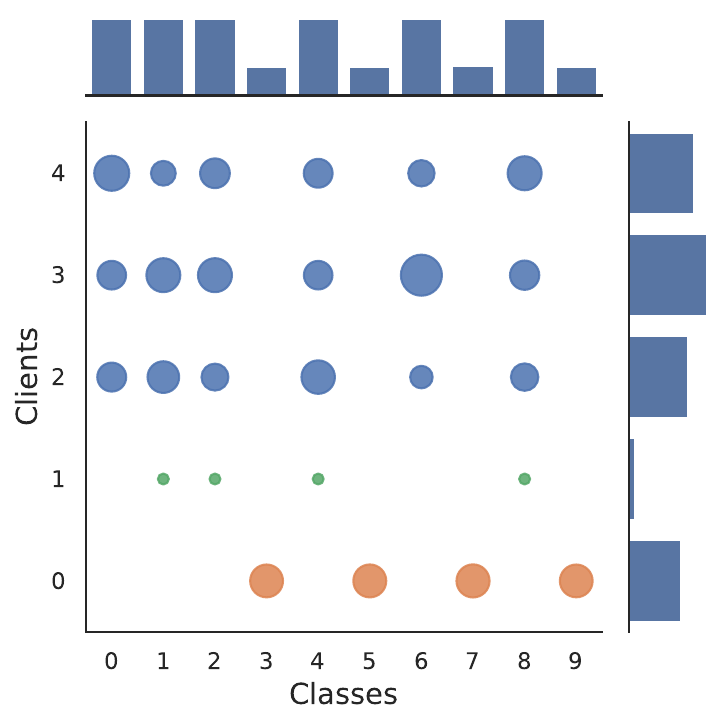}
    \vspace{2pt}
    \small (c) Maverick and Free-rider
  \end{minipage}
  \caption{Maverick versus free-rider client patterns. 
  (a) A Maverick client contributes distinctive class evidence; 
  (b) a free-rider client contributes weak or non-informative class signal; 
  (c) the mixed setting used to evaluate whether aggregation methods can identify informative rare-class clients while suppressing non-contributors.
  }
  \label{fig:maverick_free_rider}
  \vspace{-0.5em}
\end{figure*}

\vspace{-0.5em}
\noindent We next evaluate a \emph{Maverick split}, where one or more clients hold distinctive or rare classes that are weakly represented in the rest of the federation. 
This setting is important because conventional averaging can underweight such clients, even when they contain critical class information for global generalization. 
As illustrated in \cref{fig:maverick_free_rider}, Mavericks are structurally different from free-rider-like clients. 
Even when both possess few labels, a free-rider mainly carries labels that are globally well-represented. 
In contrast, Mavericks carry classes that are absent from, or are weakly represented among the other clients.
A real-world example of this setting is a specialized hospital that holds samples from rare disease categories that are weakly represented elsewhere.

\vspace{.5em}
\noindent CELM is naturally suited to this regime because its scoring rule is class-aware by construction. 
For each class, CELM computes relative class share across clients and then aggregates these class-wise shares into a final contribution score. 
As a result, a client with strong evidence on a rare class is not drowned out by clients that dominate only frequent classes. 
This behavior is what we want under Maverick splits, i.e. to reward informative rarity rather than majority overlap. 

\vspace{.5em}
\noindent \cref{tab:maverick} confirms this hypothesis quantitatively. 
On FashionMNIST, CELM gets the best balanced accuracy ($87.32$) and the strongest rare-class accuracy ($90.77$), improving substantially over baselines on rare classes. 
On CIFAR-10, CELM again leads both balanced accuracy ($68.60$) and rare-class accuracy ($61.21$), with a particularly large gain on rare classes compared to all baselines. 
Notably, CGSV collapses to near-zero rare-class performance in both datasets, highlighting the limitation of similarity-to-average scoring when rare informative updates are present. 
Overall, these results show that CELM can correctly value Maverick clients and also improve balanced and rare-class accuracy of the global model.

\begin{table*}[t]
  \centering
  \caption{Predictive performance under the Maverick split. 
  We report balanced accuracy and rare-class accuracy to assess which methods correctly value informative clients. 
  The best results are shown in \textbf{bold}. Second-best results are \underline{underlined}.\label{tab:maverick}}
  \vspace{-0.5em}
  \begin{tabular}{l l c c c c c}
    \toprule
    {\textbf{Dataset} }&  &{ \textbf{FedAvg}} & {\textbf{CFFL}} & \textbf{CGSV} &  {\textbf{ShapFed}} & \textbf{CELM} \\
    \midrule
    \multirow{2}{*}{F.MNIST} & Balanced Acc. & $84.79_{\pm 0.24}$ & $84.39_{\pm 2.43}$ & $50.74_{\pm 0.23}$ & \underline{$84.87_{\pm 0.32}$} & $\mathbf{87.32_{\pm 0.10}}$ \\
    & Rare Class Acc.   & $81.76_{\pm 0.68}$ & $82.99_{\pm 8.56}$ & $0.00_{\pm 0.00}$ & \underline{$82.02_{\pm 0.81}$} & $\mathbf{90.77_{\pm 0.24}}$ \\
    \midrule
    \multirow{2}{*}{CIFAR-10} & Balanced Acc. & \underline{$64.75_{\pm 0.29}$} & $62.10_{\pm 1.15}$ & $42.12_{\pm 4.17}$ & $64.14_{\pm 0.27}$ & $\mathbf{68.60_{\pm 0.53}}$ \\
    & Rare Class Acc.                          & \underline{$39.95_{\pm 1.06}$} & $44.69_{\pm 18.06}$ & $0.00_{\pm 0.00}$ & $38.16_{\pm 1.13}$ & $ \mathbf{61.21_{\pm 0.59}}$ \\

    \bottomrule
  \end{tabular}
\end{table*}

\begin{table*}[b!] 
  \centering
  \caption{Free-rider detection performance using z-score thresholding over contribution estimates. 
  \textbf{FR} denotes the free-rider split, and \textbf{FRM} denotes the free-rider plus Maverick split. 
  AUROC is averaged across communication rounds. 
  FPR is reported as the average false-positive rate over multiple detection thresholds. 
  Higher AUROC and lower FPR indicate better discrimination between free-rider and informative clients.\label{tab:free_rider}}
  \renewcommand{\arraystretch}{1.05}
  \setlength{\tabcolsep}{2pt}
  \begin{tabular}{l l |c c c c| c c c c}
    \toprule
    \multirow{2}{*}{{\textbf{Dataset}}} & \multirow{2}{*}{{\textbf{Split}}}
    &\multicolumn{4}{|c|}{\textbf{AUROC} ($\uparrow$)} & \multicolumn{4}{|c}{\textbf{FPR}($\downarrow$)} \\
    & & CFFL & CGSV &  ShapFed &  CELM &  CFFL & CGSV &  ShapFed &  CELM \\
    \midrule
    \multirow{2}{*}{F.MNIST}
    & FR  & 0.52  &\textbf{ 1.00}  &\textbf{ 1.00}  & \textbf{1.00}  & 0.38 & \textbf{0.00}  & 0.01  &\textbf{ 0.00} \\
    & FRM & 0.67  & 0.79  & 0.97  & \textbf{1.00}  & 0.29 & 0.24  & 0.23  & \textbf{0.08}  \\
    [0.2em]
    \midrule
    \multirow{2}{*}{CIFAR-10}
    & FR  & 0.76  & 0.96 &\textbf{ 1.00 }  &\textbf{ 1.00 } & 0.31 & 0.08  & \textbf{0.01}  & 0.03 \\
    & FRM & 0.67  & 0.08 & 0.97 & \textbf{1.00}  & 0.32  & 0.15  & \textbf{0.07}  & 0.11  \\
    [0.2em]
    \bottomrule
  \end{tabular}
  \vspace{-0.5em}
\end{table*}

\vspace{-0.5em}
\subsection{Detecting Free Riders }

\noindent Next, we evaluate free-rider detection using the synthetic client patterns shown earlier in \cref{fig:maverick_free_rider}. 
In particular, we consider two settings: (i) \textbf{FR}, where a single free-rider client is present, and (ii) \textbf{FRM}, where a free-rider co-exists with a Maverick client that carries rare but informative classes. 
This second setting is intentionally more challenging because a robust detector should suppress free-riders without mistakenly downweighting informative outlier clients.

\noindent For every communication round, we standardize client contributions into a z-score, then flag clients with z-scores below a threshold as free-riders. 
Sweeping the threshold produces a threshold--FPR curve and enables threshold-agnostic evaluation. 
\cref{tab:free_rider} reports average AUROC across rounds for FR and FRM, while the FPR numbers are computed as the average false-positive rate across multiple threshold values.

\begin{figure*}[t]
  \centering
  \begin{minipage}[t]{0.45\linewidth}
    \centering
    \includegraphics[width=\linewidth]{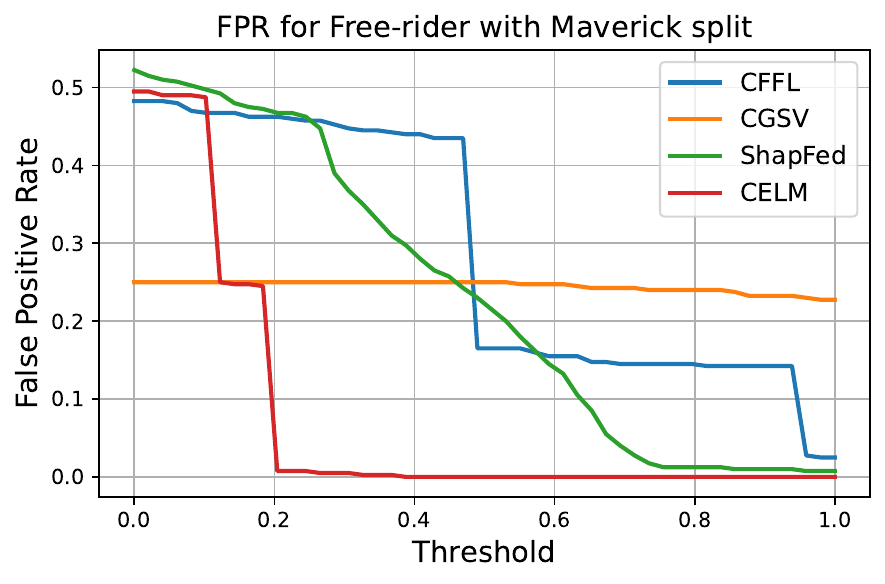}
    \small (a) FashionMNIST
  \end{minipage}\hfill
  \begin{minipage}[t]{0.45\linewidth}
    \centering
    \includegraphics[width=\linewidth]{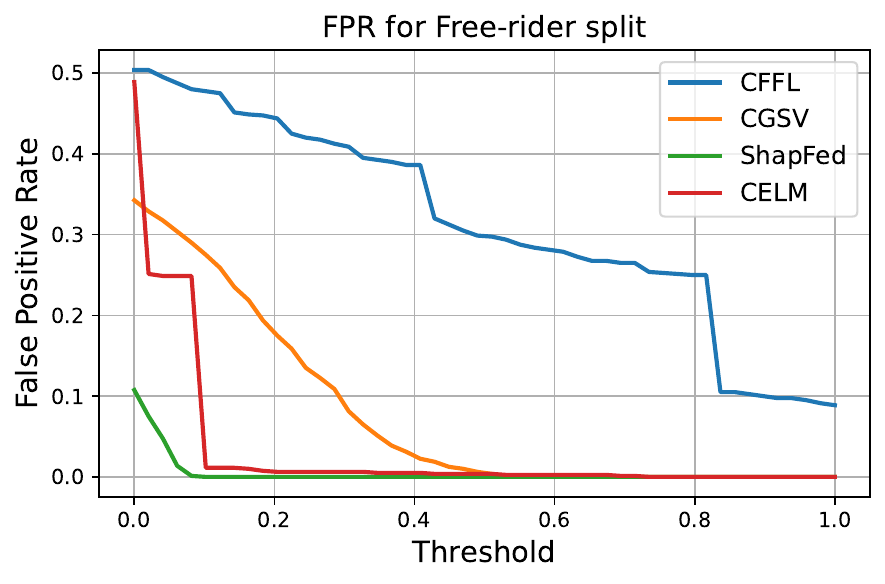}

    \small (b) CIFAR-10
  \end{minipage}

  \caption{FPR-versus-threshold curves for z-score-based free-rider detection. 
  A client is flagged as a free-rider when its standardized contribution score falls below a threshold. 
  (a) FashionMNIST under the FRM setting (free-rider + Maverick), and (b) CIFAR-10 under the FR setting (single free-rider). 
  Lower curves indicate more robust detection over a broad threshold range.}
  \label{fig:fpr}
  \vspace{-0.5em}
\end{figure*}

\noindent The results show that CELM maintains strong separability between informative and non-informative clients, especially in the more realistic FRM regime. 
\cref{fig:fpr} illustrates representative FPR-versus-threshold behavior for FashionMNIST (FRM) and CIFAR-10 (FR), where CELM remains competitive across a wide threshold range and avoids the instability seen in methods that rely on weaker class-aware signals.

\subsection{Comparison with FL Baselines for Data Heterogeneity} 

While the primary comparison in Tab.~\ref{tab:main_results} focuses on contribution-aware baselines, we include additional comparisons on selected splits against established methods in FL that mainly address data heterogeneity and label skew among clients. 
FedProx, FedNOVA, and MOON primarily modify local client objectives via regularization or representation alignment to reduce client drift from the global optimum. 
FedRS and FedUV address classifier bias induced by label imbalance. 
As shown in Tab.~\ref{tab:additional_results}, CELM performs strongly across both PLS and Maverick settings, achieving the best performance in all Maverick cases. 
FedRS is effective under PLS because using restricted softmax directly mitigates missing-label bias during local training. 

\noindent We further combine restricted softmax at the client with CELM's aggregation rule at the server to introduce the CELM (RS) variant. 
Interestingly, CELM (RS) improves over FedRS, indicating that class-wise logit maximization contributes useful server-side improvements on top of client-side label-skew corrections.
In general, 
as the modifications of the proposed method remain only at the server,
CELM can serve as a complementary improvement mechanism over any client-side optimization algorithm.

\begin{table*}[h] 
    \centering
    \caption{Additional comparison with representative FL methods for data heterogeneity. 
    We report predictive performance for Pure Label Skew (PLS) and Maverick (Mav.) settings. 
    CELM (RS) combines CELM with the restricted-softmax from FedRS.}
    \label{tab:additional_results}
        \resizebox{\linewidth}{!}{
            \begin{tabular}{l|l|ccccc|cc}
            \toprule
            Dataset  & Split & FedProx & FedNOVA & MOON & FedRS & FedUV & CELM & CELM (RS)  \\
            \midrule
            
            F.MNIST & PLS & $77.89_{\pm 1.00}$ & $80.61_{\pm 0.53}$ & $80.67_{\pm 0.29}$ & $\underline{85.37_{\pm 0.38}}$ & $82.77_{\pm 0.17}$ & $83.70_{\pm 0.19}$ & $\mathbf{85.82_{\pm 0.24}}$ \\
            F.MNIST & Mav. & $80.95_{\pm 0.10}$ & $84.89_{\pm 0.21}$ & $82.68_{\pm 0.48}$ & $85.70_{\pm 0.38}$ & $84.79_{\pm 0.23}$ & $\mathbf{87.32_{\pm 0.10}}$ & $\underline{85.96_{\pm 0.33}}$ \\
            \midrule
            CIFAR-10& PLS  & $50.15_{\pm 3.27}$ & $55.21_{\pm 3.63}$ & $53.93_{\pm 2.81}$ & $\underline{62.90_{\pm 0.50}}$ & $49.65_{\pm 2.71}$  & $59.11_{\pm 1.96}$ & $\mathbf{64.05_{\pm 0.54}}$ \\
            CIFAR-10 & Mav. & $58.83_{\pm 0.45}$ & $\underline{64.90_{\pm 0.28}}$ & $59.07_{\pm 1.29}$ & $62.49_{\pm 0.80}$ & $57.02_{\pm 1.69}$ & $\mathbf{68.60_{\pm 0.53}}$ & $63.74_{\pm 0.93}$\\
            \bottomrule
            \end{tabular}
        }
        \vspace{-1em}
\end{table*}

\vspace{-0.8em}
\subsection{Estimation Fidelity}
\begin{figure*}[t]
  \centering
  \begin{minipage}[t]{0.33\linewidth}
    \centering
    \includegraphics[width=\linewidth]{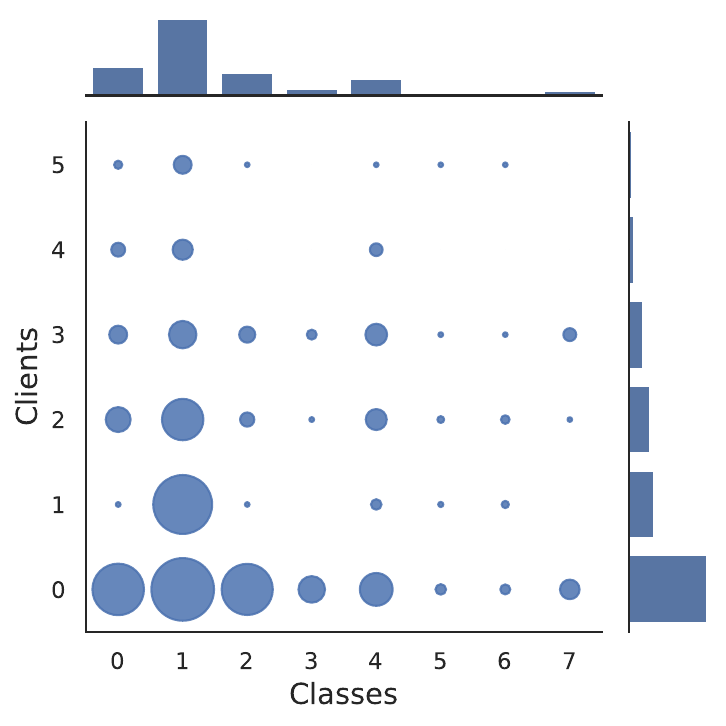}
    \small (a) True Distribution
  \end{minipage}\hspace{25pt}
  \begin{minipage}[t]{0.46\linewidth}
    \centering
    \includegraphics[width=\linewidth]{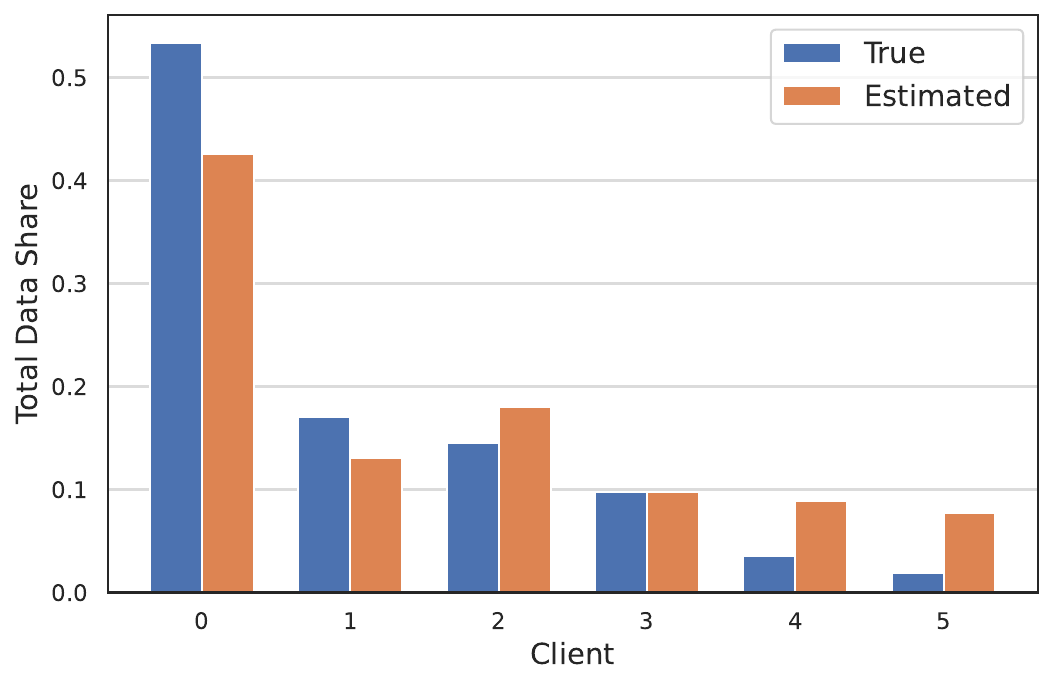}
    \small (c) Client Marginals
  \end{minipage}

  \vspace{3pt}

  \begin{minipage}[t]{0.33\linewidth}
    \centering
    \includegraphics[width=\linewidth]{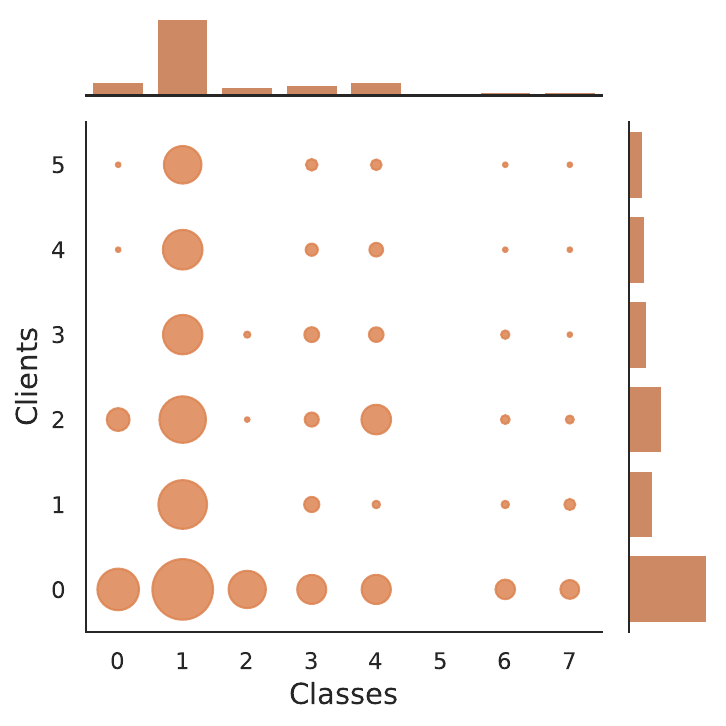}
    \small (b) Estimated Distribution
  \end{minipage}\hspace{25pt}
  \begin{minipage}[t]{0.46\linewidth}
    \centering
    \includegraphics[width=\linewidth]{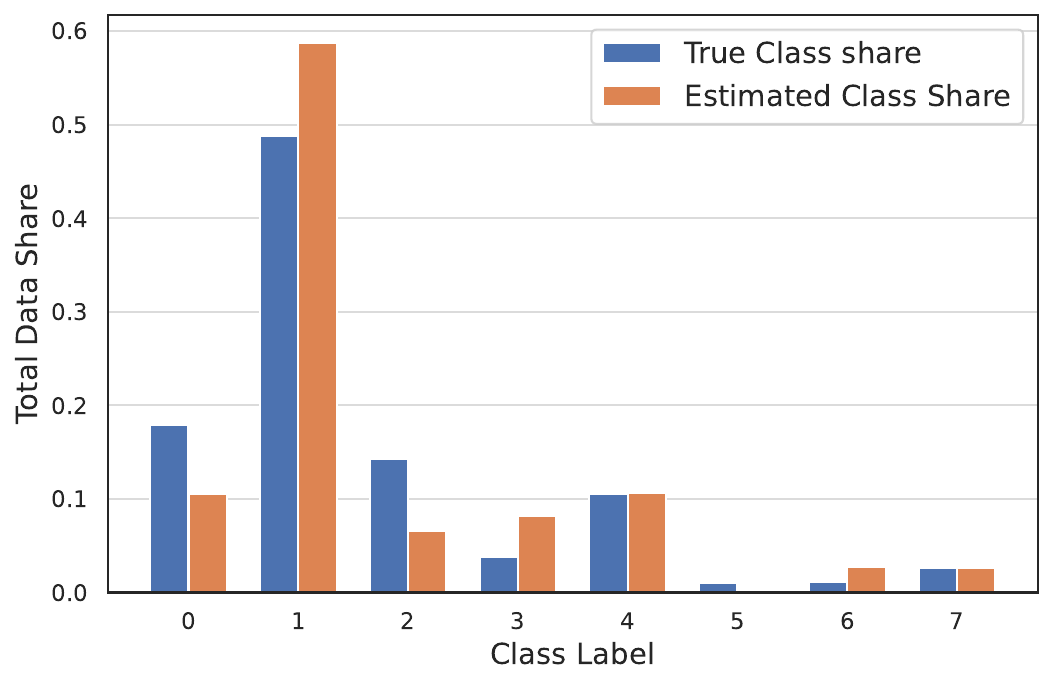}

    \small (d) Global Class Marginals
  \end{minipage}
  \caption{Estimation fidelity of CELM on FedISIC dataset. Panels (a) and (b) compare the true and CELM-estimated client class distributions, while panels (c) and (d) compare client-level marginals and the aggregated global class marginals. 
  The close visual agreement indicates that CELM preserves both per-client class tendencies and cohort-level class prevalence without accessing raw client data.}
  \label{fig:est_fidelity_grid}
  \vspace{-1em} 
\end{figure*}

\vspace{-0.2em}
\noindent \cref{fig:est_fidelity_grid} and \cref{tab:fidelity} jointly evaluate how accurately CELM recovers client-level and cohort-level class structure using the class-wise evidence ($q_{i,c}$) extracted through logit-maximization.
\cref{fig:est_fidelity_grid} shows a qualitative view. Panel (a) shows the true client distribution, panel (b) shows the CELM-estimated distribution, and panels (c)–(d) compare client marginals and global class marginals, respectively. 
Visually, the estimated distributions track the dominant classes and relative mass allocation patterns in the true distributions, indicating that CELM captures not only marginal trends but also global client-class distribution heterogeneity.

\vspace{.5em}
\noindent\cref{tab:fidelity} complements this visual evidence with quantitative distribution-distance metrics between true and estimated distributions. 
We report Jensen--Shannon Divergence (JSD) \cite{manning1999foundations}, Earth Mover's Distance (EMD) \cite{rubner1998metric}, and Hellinger distance \cite{nikulin2001hellinger}, and include a uniform-distribution baseline for context. 
Across all reported datasets/splits, CELM yields lower distances than the uniform baseline, showing that CELM estimates are closer to the true data distributions.

\begin{table*}[t!]

  \centering
  \caption{Distribution-modelling fidelity of CELM. We report distances between the true class distribution and either (i) a uniform reference distribution or (ii) the CELM-estimated distribution. Lower is better ($\downarrow$). 
  JSD: Jensen-Shannon Divergence, EMD: Earth Mover's Distance, and Hellinger: Hellinger distance. 
  \label{tab:fidelity}}
\setlength{\tabcolsep}{3pt}
  \begin{tabular}{l l |c c |c c | c c}

    \toprule

    \multirow{2}{*}{{\textbf{Dataset}}} & \multirow{2}{*}{{\textbf{Split}}}
      &\multicolumn{2}{|c|}{\textbf{JSD} ($\downarrow$)} & \multicolumn{2}{|c|}{\textbf{EMD}($\downarrow$)} & \multicolumn{2}{|c}{\textbf{Hellinger}($\downarrow$)} \\
      & & Uniform & CELM & Uniform &  CELM & Uniform & CELM \\
    \midrule

    \multirow{2}{*}{F.MNIST}
 & Dir. (0.05) & 0.613 & 0.294 & 0.114 & 0.079 & 0.693 & 0.306 \\
 & PLS & 0.508 & 0.231 & 0.077 & 0.053 & 0.586 & 0.241 \\
 [0.2em]
    \midrule
    \multirow{2}{*}{CIFAR-10}
    & Dir. (0.01) & 0.646 & 0.145 & 0.124 & 0.036 & 0.735 & 0.151 \\
 & SLS & 0.433 & 0.200 & 0.066 & 0.046 & 0.500 & 0.201 \\
 [0.2em]
    \midrule
    FedISIC & Natural & 0.535 & 0.265 & 0.097 & 0.051 & 0.583 & 0.286  \\
    \bottomrule
  \end{tabular}
\end{table*}

\vspace{-1em}
\section{Conclusion \& Limitations}
\vspace{-0.5em}
\noindent Our results consistently show that class-wise data-free contribution estimation can improve federated aggregation under realistic non-IID conditions. 
Across synthetic and natural splits, CELM shows strong gains in both overall and minority-sensitive metrics, indicating that class-aware weighting is more robust than purely size-based or similarity-to-average aggregation. 
The Maverick and free-rider experiments further strengthen the conclusion that CELM not only improves global predictive performance but also better distinguishes informative outlier clients from non-contributing clients. 
Together with the estimation-fidelity analyses, these results suggest that logit probing captures a meaningful distributional structure that is directly useful for server-side decision making.

\noindent At the same time, CELM introduces additional server-side compute during warm-up due to repeated logit-maximization probes across clients and classes.
Because of this overhead, our experiments remain in the small-to-medium-scale model regime typical of controlled FL evaluation.
Although this overhead is bounded by the freeze-after-warm-up design, scaling CELM to substantially larger numbers of clients, classes, and high-resolution tasks is an important direction that may require subsampling or adaptive class selection. 
We discuss the compute complexity of our approach further in Appendix \ref{app:compute}.
Another practical consideration is that detection quality depends on model confidence calibration. Highly miscalibrated client models can distort relative evidence estimates. 
Future work can therefore explore calibration-aware probing, adaptive warm-up schedules, and theoretical guarantees on contribution identifiability under extreme heterogeneity and adversarial participation.
\vspace{-1.5em} 
\subsection*{Acknowledgements} 
This material is partly based on work supported by the Office of Naval Research N00014-24-1-2168. 

\bibliographystyle{splncs04}
\bibliography{main}

\clearpage

\appendix 

\section{Algorithm}

\begin{algorithm}[H]
  \caption{\texttt{CELM}: Contribution Estimation from Logit Maximization}
  \label{alg:celm}
  \renewcommand{\algorithmicrequire}{\textbf{Input:}}
  \renewcommand{\algorithmicensure}{\textbf{Procedure}}
  \begin{algorithmic}[1]
    \Require Initial global model $w_g^{(0)}$, clients $\mathcal{N}=\{1,\dots,N\}$, classes $\mathcal{Y}=\{1,\dots,K\}$, total rounds $T$, warm-up rounds $T_w$, LM steps $L$, LM step size $\eta$, regularization weight $\lambda$, EMA factor $\beta$

    \Function{LMProbe}{$w,\mathbf{X}^{\mathrm{init}},\mathcal{Y},L,\eta,\lambda$}

    \ForAll{class $c\in\mathcal{Y}$ \textbf{in parallel}}
    \State $\mathbf{x}_{c}^{(0)} \gets \mathbf{X}^{\mathrm{init}}_c$
    \For{$\ell=0$ to $L-1$}
    \State $g_c^{(\ell)} \gets \nabla_\mathbf{x}\big(s_{c}(\mathbf{x}_{c}^{(\ell)}; w)-\lambda\lVert \mathbf{x}_{c}^{(\ell)}\rVert_2^2\big)$
    \State $\mathbf{x}_{c}^{(\ell+1)} \gets \Call{Adam}{\mathbf{x}_{c}^{(\ell)}, g_c^{(\ell)}, \eta, \beta_1=0.9, \beta_2=0.999}$
    \EndFor
    \State $u_c \gets s_{c}(\mathbf{x}_{c}^{(L)}; w),\; \mathbf{X}^{\mathrm{final}}_c \gets \mathbf{x}_{c}^{(L)}$
    \EndFor
    \State \Return $\{u_c\}_{c=1}^{K},\mathbf{X}^{\mathrm{final}}$
    \EndFunction
    \State Initialize $c_i^{(0)} \gets 1/N,\; \forall ~i\in\mathcal{N}$
    \State Initialize $\mathbf{X}_{g}^{(0)}$ and $\mathbf{X}_{i}^{(0)} \;\forall i\in\mathcal{N}$ from Standard Gaussian Noise

    \For{$t=1$ to $T$}
    \If{$t \le T_w$}
    \State Broadcast global backbone $w_g^{(t-1)}$; clients retain local heads
    \Else
    \State Broadcast full global model $w_g^{(t-1)}$
    \EndIf
    \State Clients perform local training and return $\{w_i^{(t)}\}_{i=1}^{N}$
    \If{$t \le T_w$}
    \State $\{u_{g,c}^{(t)}\}_{c=1}^{K},\mathbf{X}_{g}^{(t)} \gets \Call{LMProbe}{w_g^{(t-1)},\mathbf{X}_{g}^{(t-1)},\mathcal{Y},L,\eta,\lambda}$
    \State $b^{(t-1)} \gets \frac{1}{K}\sum_{c=1}^{K} u_{g,c}^{(t)}$
    \For{each client $i\in\mathcal{N}$}
    \State $\{u_{i,c}^{(t)}\}_{c=1}^{K},\mathbf{X}_{i}^{(t)} \gets \Call{LMProbe}{w_i^{(t)},\mathbf{X}_{i}^{(t-1)},\mathcal{Y},L,\eta,\lambda}$
    \For{each class $c\in\mathcal{Y}$}

    \State $q_{i,c}^{(t)} \gets \max\big(0,{u}_{i,c}^{(t)}-b^{(t-1)}\big)$
    \EndFor
    \EndFor
    \State Compute $r_{i,c}^{(t)} \gets q_{i,c}^{(t)} /((\sum_{j=1}^{N} q_{j,c}^{(t)}) +\epsilon)$
    \State Compute $\hat{c}_i^{(t)} \gets \frac{1}{K}\sum_{c=1}^{K} r_{i,c}^{(t)}$
    \State Instantaneous score: $\bar{c}_i^{(t)} \gets \hat{c}_i^{(t)} / \sum_{j=1}^{N} \hat{c}_j^{(t)}$
    \State EMA smoothing: $c_i^{(t)} \gets \beta c_i^{(t-1)} + (1-\beta)\bar{c}_i^{(t)}$
    \Else
    \State Freeze weights: $c_i^{(t)} \gets c_i^{(T_w)}$
    \EndIf
    \State Aggregate global model: $w_g^{(t)} \gets \sum_{i=1}^{N} c_i^{(t)} w_i^{(t)}$
    \EndFor
  \end{algorithmic}
\end{algorithm}

\section{Additional Results}
\subsection{Results on Homogeneous Partition}

\noindent \cref{tab:iid_results} reports performance under homogeneous (IID) client distributions. 
We simulate homogeneous behavior by setting a high value for the skew parameter ($\alpha = 100$) for the Dirichlet split. 
In this regime, all methods operate near their upper-bound behavior because there is little cross-client distribution shift to correct. 
CELM remains competitive with all the baselines on both datasets, indicating that class-wise contribution weighting does not introduce a penalty when client data are already balanced.

\begin{table*}[h]

  \centering
  \caption{
  Global predictive performance under homogeneous (IID) client distributions. 
  We report test accuracy (\%,$\text{mean}_{\pm \text{std.}}$) on FashionMNIST and CIFAR-10. 
  This setting serves as a sanity check showing that CELM remains competitive when non-IID skew is minimal. \label{tab:iid_results}
  }

  \begin{tabular}{l l c c c c c}
    \toprule

    {\textbf{Dataset} }& {\textbf{Split}} &{ \textbf{FedAvg}} & {\textbf{CFFL}} & \textbf{CGSV} &  {\textbf{ShapFed}} & \textbf{CELM} \\

    \midrule
    F.MNIST & IID & $89.23_{\pm 0.09}$ & $89.04_{\pm 0.04}$ & $86.57_{\pm 0.25}$ & $89.19_{\pm 0.07}$ & $89.16_{\pm 0.05}$ \\
    \midrule
    CIFAR-10 & IID & $76.76_{\pm 0.46}$ & $76.40_{\pm 0.75}$ & $62.65_{\pm 1.41}$ & $76.51_{\pm 0.52}$ & $76.15_{\pm 0.47}$ \\
    \bottomrule
  \end{tabular}
  \vspace{-3em}
\end{table*}

\subsection{Number of Clients}

\noindent \cref{tab:large_nclients} studies scalability of CELM to larger number of clients ($N= 20$) under Dirichlet non-IID splits. 
As the number of clients increases, each client typically sees fewer samples and stronger local skew, making contribution estimation harder. 
CELM is consistently best or tied-best across the splits, with the largest gains on CIFAR-10 under stronger heterogeneity (Dirichlet, $\alpha=0.01$ ). 

\begin{table*}[h]

  \centering
  \caption{
  Global predictive performance for 20 clients under non-IID partitions. 
  We report test accuracy (\%,$\text{mean}_{\pm \text{std.}}$); best values per row are shown in \textbf{bold}. 
  CELM shows consistent improvements, especially under stronger skew. \label{tab:large_nclients}
  }

  \begin{tabular}{l l c c c c c}
    \toprule

    {\textbf{Dataset} }& {\textbf{Split}} &{ \textbf{FedAvg}} & {\textbf{CFFL}} & \textbf{CGSV} &  {\textbf{ShapFed}} & \textbf{CELM} \\

    \midrule
    \multirow{3.2}{*}{F.MNIST}
    & Dir. (0.01) & $77.20_{\pm 1.41}$ & $71.76_{\pm 1.61}$ & $24.48_{\pm 6.74}$ & $77.29_{\pm 1.48}$ & $\mathbf{78.71_{\pm 1.22}}$ \\
    & Dir. (0.05) & $80.28_{\pm 1.05}$ & $78.76_{\pm 2.36}$ & $45.50_{\pm 7.84}$ & $80.39_{\pm 1.17}$ & $\mathbf{81.70_{\pm 1.05}}$ \\
    & Dir. (0.1) & $82.81_{\pm 1.06}$ & $77.51_{\pm 3.81}$ & $59.22_{\pm 3.66}$ & $\mathbf{82.88_{\pm 0.96}}$ & $\mathbf{82.88_{\pm 1.27}}$ \\
     [0.3em]
    \midrule
    \multirow{3.2}{*}{CIFAR-10}
    & Dir. (0.01) & $36.17_{\pm 4.36}$ & $28.57_{\pm 5.63}$ & $15.47_{\pm 2.51}$ & $36.62_{\pm 4.79}$ & $\mathbf{44.85_{\pm 1.78}}$ \\
    & Dir. (0.05) & $48.33_{\pm 3.28}$ & $39.57_{\pm 8.49}$ & $25.09_{\pm 4.29}$ & $48.65_{\pm 3.34}$ & $\mathbf{51.00_{\pm 2.96}}$ \\
    & Dir. (0.1) & $53.91_{\pm 1.62}$ & $51.03_{\pm 4.61}$ & $28.38_{\pm 3.08}$ & $54.05_{\pm 1.84}$ & $\mathbf{54.58_{\pm 2.16}}$ \\
    \bottomrule
  \end{tabular}
  \vspace{-2em}
\end{table*}

\subsection{Effect of number of classes} 

\noindent \cref{tab:large_classes} compares the performance of CELM on the EMNIST dataset with a larger number of labels (47 classes), across multiple non-IID partitioning schemes. 
We simulate this split with 5 clients. Compared with the lower-class-count benchmarks, this setting is more sensitive to noisy client evidence. 
CELM remains competitive across most splits, albeit with smaller improvement margins than in the low-label-count setting.

\begin{table*}[h]
\vspace{-1em}

  \centering
  \caption{
  Global predictive performance on a larger-class benchmark (EMNIST) across Dirichlet, PLS, and SLS non-IID splits. 
  We report test accuracy (\%,$\text{mean}_{\pm \text{std.}}$), with best values per row in \textbf{bold}. 
  \label{tab:large_classes}}

  \begin{tabular}{l l c c c c c}
    \toprule
    {\textbf{Dataset} }& {\textbf{Split}} &{ \textbf{FedAvg}} & {\textbf{CFFL}} & \textbf{CGSV} &  {\textbf{ShapFed}} & \textbf{CELM} \\
    \midrule
    \multirow{5.2}{*}{EMNIST}
    & Dir. (0.01) & $82.12_{\pm 0.15}$ & $81.67_{\pm 0.63}$ & $15.02_{\pm 2.89}$ & $82.23_{\pm 0.20}$ & $\mathbf{82.42_{\pm 0.32}}$ \\
    & Dir. (0.05) & $82.99_{\pm 0.77}$ & $83.09_{\pm 0.35}$ & $22.12_{\pm 7.11}$ & $82.99_{\pm 0.79}$ & $\mathbf{83.15_{\pm 0.49}}$ \\
    & Dir. (0.1) & $84.37_{\pm 0.44}$ & $\mathbf{85.00_{\pm 0.12}}$ & $33.89_{\pm 3.31}$ & $84.45_{\pm 0.49}$ & $84.35_{\pm 0.28}$ \\
    & PLS & $82.02_{\pm 0.92}$ & $81.98_{\pm 0.93}$ & $35.74_{\pm 1.72}$ & $82.16_{\pm 0.85}$ & $\mathbf{82.41_{\pm 0.89}}$ \\
    & SLS & $83.47_{\pm 0.34}$ & $\mathbf{86.18_{\pm 0.45}}$ & $31.34_{\pm 14.28}$ & $83.91_{\pm 0.37}$ & $85.25_{\pm 0.26}$ \\
    \bottomrule
  \end{tabular}
  \vspace{-2em}
\end{table*}
\vspace{-0.5em}

\section{Ablation study}
\label{app:ablation}

\subsection{Warm-up rounds}

\begin{table*}[b]
  \vspace{-1em}
  \centering
  \caption{
  Warm-up horizon ablation for CELM. We report test accuracy (\%,$\text{mean}_{\pm \text{std.}}$) for different warm-up fractions $T_w/T$. 
  Smaller $T_w$ reduces probing cost, while larger $T_w$ can help in better estimation of the class-evidence matrix.
  \label{tab:tw_ablation}
  }

  \begin{tabular}{l l | c c c c c}
    \toprule
   \multirow{2}{*}{{\textbf{Dataset}}} & \multirow{2}{*}{{\textbf{Split}}}
    &\multicolumn{5}{|c}{\textbf{$T_w$ as \% of total rounds $T$}} \\
    & & 5 \% & 10 \% &  15 \% &  20 \% &  30 \% \\

    \midrule
    \multirow{2.2}{*}{F.MNIST}
    & Dir. (0.05) & $83.64_{\pm 0.42}$ & $83.31_{\pm 0.99}$ & $83.07_{\pm 0.95}$ & $83.06_{\pm 0.75}$ & $82.70_{\pm 0.91}$ \\
    & PLS         & $83.70_{\pm 0.19}$ & $83.57_{\pm 0.45}$ & $83.73_{\pm 0.34}$ & $83.99_{\pm 0.10}$ & $84.14_{\pm 0.51}$ \\
     [0.3em]
    \midrule
    \multirow{2.2}{*}{CIFAR-10}
    & Dir. (0.05) & $67.16_{\pm 1.29}$ & $66.88_{\pm 1.54}$ & $66.68_{\pm 1.49}$ & $66.54_{\pm 1.43}$ & $66.43_{\pm 1.49}$ \\
    & PLS         & $59.11_{\pm 1.96}$ & $59.03_{\pm 2.32}$ & $59.36_{\pm 1.42}$ & $59.44_{\pm 0.95}$ & $59.86_{\pm 0.57}$ \\
    \bottomrule
  \end{tabular}
  \vspace{-1em}
\end{table*}

\noindent \cref{tab:tw_ablation} studies the warm-up horizon $T_w$ as a fraction of total communication rounds. 
Choosing $T_w$ creates a trade-off between estimation quality and efficiency. 
A longer warm-up provides more rounds to estimate class-wise evidence before freezing contributions, but it also increases server-side probing costs and can slow global convergence because classifier heads are not shared during this stage. 
A shorter warm-up reduces computation and speeds up full-model synchronization, but it may produce noisier estimates of the class-evidence matrix ($Q$), limiting the benefit of contribution-aware weighting. 
This trend is reflected in the results: for Dirichlet($\alpha=0.05$), a shorter warm-up performs best and a longer warm-up leads to saturation or mild decline; 
for Pure Label Skew, a longer warm-up improves performance, consistent with stronger class asymmetry requiring more evidence collection. 
To balance accuracy and computational overhead, we set $T_w=0.05T$ in the main experiments. 

\begin{figure*}[t]
  \centering
  \includegraphics[width=\linewidth]{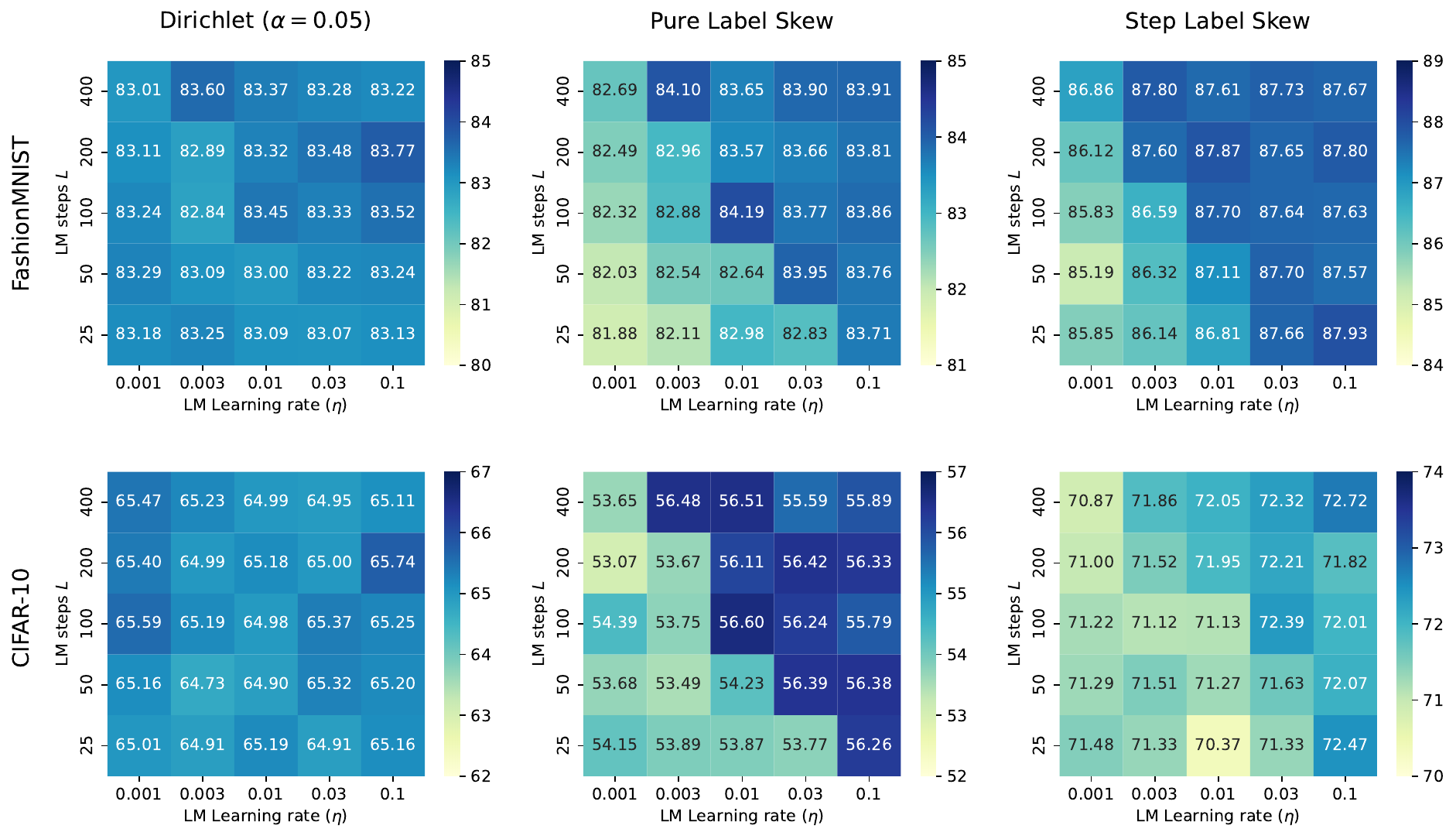}
  \caption{Sensitivity of CELM to LM optimization steps $L$ and LM learning rate $\eta$. 
  Each heatmap cell reports final test accuracy (\%) for a specific $(L,\eta)$ pair across representative non-IID splits on FashionMNIST and CIFAR-10.
  }
  \label{fig:lm_ablation}
  \vspace{-1.5em} 
\end{figure*}
\subsection{Ablation on LM steps  and LM learning rate}
\noindent \cref{fig:lm_ablation} presents a grid search over LM optimization steps $L$ and LM learning rate $\eta$ on representative non-IID splits of FashionMNIST and CIFAR-10. 
Very small learning rates (e.g., $0.001$) and shallow probes (e.g., $L=25$) under-optimize class evidence, while very large $L$ provides diminishing returns. 
Across settings, a broad high-performing region appears for $\eta\in[0.01,0.1]$ and $L\in[100,400]$, indicating that CELM is reasonably robust to moderate hyperparameter variation. 
In the main experiments, we use $\eta=0.01$ and $L=200$ as a stable operating point that balances accuracy and probing cost. 
When compute is constrained, higher-$\eta$, smaller-$L$ configurations are viable alternatives.

\vspace{-0.5em} 
\section{Compute Complexity}
\label{app:compute}

\noindent Let $N$ denote the number of clients, $K$ the number of classes, $T_w$ the warm-up rounds, and $L$ the LM optimization steps. 
Let $C_{\mathrm{LM}}$ be the cost of one forward-backward Logit Maximization step with respect to the input for a single class and model. 
This cost depends on the model size and the input image dimensions. 
The per-model compute cost for running all LM steps is  $\mathcal{O}\big(K\,L\,C_{\mathrm{LM}}\big)$. 
Importantly, the per-class LM loop is embarrassingly parallel. In our implementation, we optimize all classes simultaneously. 
The per-model latency then becomes $\mathcal{O}\big(L\,C_{\mathrm{LM}}\big)$.
CELM runs LM on the global model and all client models for $T_w$ rounds, leading to a total server-side complexity of $\mathcal{O}\big(T_w\,(N+1)\,L\,C_{\mathrm{LM}}\big)$.

\noindent The additional memory overhead is dominated by cached warm-start initial images and class-wise evidence tensors, i.e., $\mathcal{O}((N+1)K\lvert\mathbf{x}\rvert + NK) \sim \mathcal{O}(NK\lvert\mathbf{x}\rvert) $. 
After warm-up ($t>T_w$), contribution weights are frozen, and CELM reduces to standard weighted aggregation with cost comparable to FedAvg, while communication overhead remains unchanged. 
Since LM is independent across models as well, there is further scope for reducing latency at the cost of additional memory.

\section{Optimized LM image samples}
\begin{figure*}[h]
\vspace{-2.5em}
  \centering
  \includegraphics[width=\linewidth]{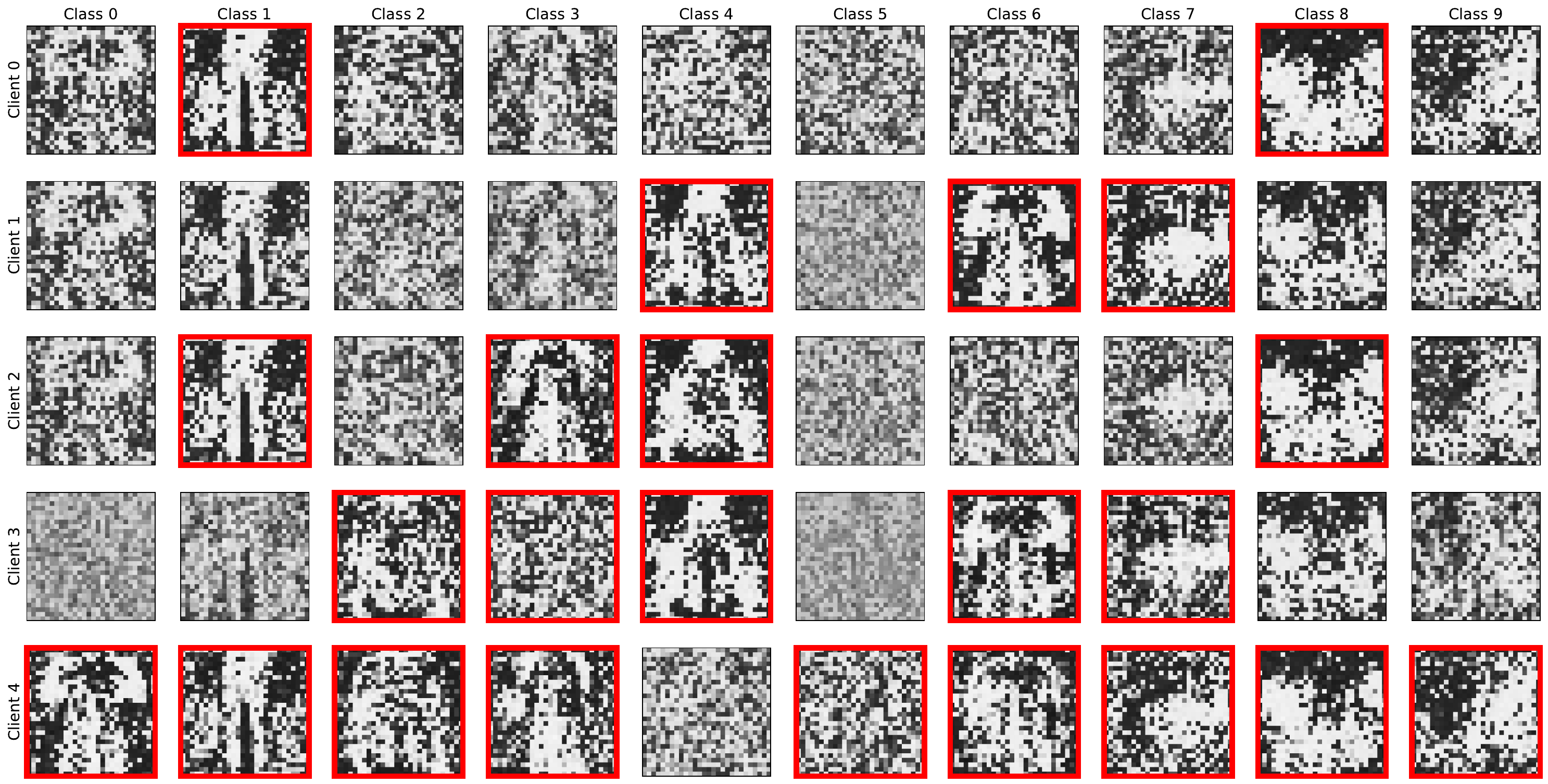}
  \caption{Final warm-up LM probe images ($\mathbf{X}^{\mathrm{final}}$) on FashionMNIST (Pure Label Skew). 
  Each row is a client, and each column is a target class; red outlines indicate classes present in that client's local data. 
  Present-class cells often exhibit more distinct response patterns, while the images remain abstract and non-semantic, indicating useful class evidence without visually revealing recognizable training samples.}
  \label{fig:lm_outputs}
  \vspace{-1em} 
\end{figure*}
\noindent \cref{fig:lm_outputs} provides a qualitative view of the final LM images used by CELM after the warm-up phase. 
Rows correspond to clients and columns correspond to target classes; red outlines mark classes present in a client's local data. 
In most cases, cells aligned with present classes show clearer and more distinct activation patterns with higher contrasts than absent-class cells, supporting the role of LM as class-selective evidence for contribution estimation.

\noindent At the same time, these LM outputs remain highly synthetic and do not show semantic value. 
They do not resemble recognizable training samples and do not reveal client-specific content in a human-interpretable way. 
This is expected because CELM optimizes random Gaussian initializations for class-logit response (with only $\ell_2$ regularization), and not for data reconstruction. 
Therefore, while LM outputs are useful for relative class evidence scoring, we do not observe direct visual leakage in these probes.

\end{document}